\documentclass[conference]{IEEEtran}
\usepackage{fix-cm}
\usepackage[utf8]{inputenc}
\usepackage{cite}
\usepackage{amsmath,amssymb,amsfonts}
\usepackage{algorithmic}
\usepackage{graphicx}
\usepackage{textcomp}
\usepackage{xcolor}
\usepackage{url}
\usepackage{xurl}
\usepackage{tabularx}
\usepackage{tikz}
\usetikzlibrary{patterns}
\usepackage{threeparttable}
\usepackage{array}
\usepackage{hyperref}
\usepackage{booktabs}
\definecolor{BaseGray}{HTML}{5A5A5A}
\usepackage{pgfplots}
\usepgfplotslibrary{groupplots}

\pgfplotsset{compat=1.18}
\usepackage{colortbl}
\usepackage[inline]{enumitem}
\usepackage{ragged2e} 
\usepackage{tabularx}

\definecolor{OIblue}{HTML}{0072B2}
\definecolor{OIorange}{HTML}{E69F00}
\definecolor{BaseGray}{HTML}{666666}
\definecolor{LightBlue}{HTML}{E6F2FA}
\definecolor{LightOrange}{HTML}{FFF1D6}
\definecolor{LightGray}{HTML}{F2F2F2}
\definecolor{PanelGray}{HTML}{EDEDED}
\definecolor{HeaderGray}{HTML}{F7F7F7}
\definecolor{BasePurple}{HTML}{7A3E9D}

\def\BibTeX{{\rm B\kern-.05em{\sc i\kern-.025em b}\kern-.08em
    T\kern-.1667em\lower.7ex\hbox{E}\kern-.125emX}}
\begin{document}

\title{LuxSQA: Ask Me in Luxembourgish with TTS-Augmented Spoken Question Answering}

\author{
\IEEEauthorblockN{1\textsuperscript{st} Nina Hosseini-Kivanani}
\IEEEauthorblockA{
\textit{Radio T{\'e}l{\'e}visioun L{\"e}tzebuerg (RTL)}\\
\textit{University of Luxembourg}\\
Esch-sur-Alzette, Luxembourg\\
nina.hosseinikivanani@ext.uni.lu
}
\and
\IEEEauthorblockN{2\textsuperscript{nd} Marco Matassoni}
\IEEEauthorblockA{
\textit{Fondazione Bruno Kessler (FBK)}\\
Trento, Italy\\
matasso@fbk.eu
}
\and
\IEEEauthorblockN{3\textsuperscript{rd} Alessio Brutti}
\IEEEauthorblockA{
\textit{Fondazione Bruno Kessler (FBK)}\\
Trento, Italy\\
brutti@fbk.eu
}
}


\maketitle

\begin{abstract}

Spoken Question Answering (SQA) remains largely focused on high-resource languages and carefully recorded speech, limiting the reach of speech-LLM methods in low-resource settings. This paper investigates whether text-to-speech (TTS) can provide task-specific training data for Luxembourgish SQA without requiring a large human-recorded QA corpus. Starting from existing text-based QA resources, we translate questions into Luxembourgish, synthesize spoken questions with multiple TTS systems, and pair them with textual answers. We train a parameter-efficient SLAM-style architecture that connects a frozen Whisper encoder to frozen multilingual LLM backends through a learned projector and LoRA adapters. We compare MMS-TTS, Qwen3-TTS, and OmniVoice variants, including single-source corpora of about 48k questions and a 4TTS multi-source mix of approximately 230k questions. Evaluation on LLAMA-LB-Test with two real Luxembourgish speaker conditions shows that multi-source and voice-design-based synthetic training configurations yield the strongest SQA performance. The results also show that no-reference TTS quality scores do not monotonically predict downstream QA performance, indicating that synthetic speech must be evaluated as task-specific training data rather than only as natural-sounding audio.

\end{abstract}


\section{Introduction}

Spoken Question Answering (SQA) extends question answering to spoken inputs, requiring models to understand both the linguistic content and the acoustic characteristics of speech~\cite{you-etal-2022-end}. Therefore, SQA must cope with disfluencies, informal phrasing, speaker and channel variability, and background noise. Early systems mainly used a cascade approach, where an automatic speech recognition (ASR) module transcribed the audio before a text QA model produced the answer. Although effective in clean conditions, this design suffers from error propagation when the ASR output is noisy, and it discards prosodic cues such as prominence and phrasing that can help localize answer spans~\cite{sidiropoulos2022impact}.


Recent speech-LLM architectures address these issues by coupling powerful speech encoders with instruction-tuned language models via a lightweight connector or projector: the projector maps acoustic embeddings into the token/embedding space of a frozen LLM so that only a small number of parameters need updating. This parameter-efficient connector idea (small adapters + LoRA) has been shown to yield strong performance while keeping compute and data requirements modest (e.g., Speech, Language, Audio and Music-ASR (SLAM-ASR) and related LLM-ASR studies); ~\cite{wang2023slm,ma2024slam,ma2025speech,ma2026slam} illustrate the framework.
Follow-up work has explored robustness, scaling, and multilingual transfer of these projector-based systems; see evaluations and references in recent SLAM/SLM/SLAM-ASR analyses~\cite{kumar2025performance}. The projector + LLM approach is attractive for low-resource scenarios because it reuses large pretrained components and pushes most task adaptation into a compact layer that can be trained with limited hardware and data; several articles explicitly study data volume, multilingual transfer, and adapter-based solutions for low-resource speech tasks~\cite{fong2025speech}.

This raises a practical question: Can we leverage TTS to build useful SQA training data without recording new speakers, and if so, how sensitive is the resulting system to TTS quality, voice diversity, and corpus size? Previous work that automatically generates SQA data or converts text QA corpora to speech shows that this is a feasible and useful strategy: frameworks for automatic SQA generation~\cite{you2020towards} and experiments with synthetic speech have been proposed and evaluated, with explicit analyzes of the choice of TTS and ASR noise effects. See the automatic generation framework and follow-ups by Menevşe et al.~\cite{menevcse2024dealing,menevse2022framework} and empirical dataset studies such as HeySQuAD and LibriSQA~\cite{wu2023heysquad,zhao2024librisqa}. 

Early SQA systems mainly relied on ASR followed by text-based question answering, making performance strongly dependent on transcription quality. More recent work has moved toward direct modeling of spoken inputs. End-to-end SQA and spoken conversational QA have been studied for multi-turn dialogue and speech-based answer generation~\cite{you-etal-2022-end,chen2021self,chen-etal-2025-slam}. Generative systems such as GSQA reduce reliance on explicit ASR transcripts by mapping spoken questions to spoken or textual answers, while Spectron adapts pretrained LLMs to speech input and output using spectrogram-based representations~\cite{shih2024gsqa,nachmani2024spoken}. These studies show that direct SQA is feasible, but they usually require substantial task-specific training and do not focus on low-resource languages. A related line of work connects speech encoders to LLMs through lightweight projectors. LLM-based ASR and SLAM-style systems map acoustic embeddings into the LLM embedding space and can achieve competitive performance while keeping the speech encoder and LLM frozen or lightly tuned~\cite{fong2025speech,ma2024slam}. Instruction-following speech LLMs extend this approach to tasks such as spoken summarization, classification, and dialogue, including benchmarks with closed-ended and open-ended speech instructions~\cite{wang2025inserter}. Most of this work, however, targets transcription or general speech instruction following rather than low-resource SQA. Synthetic speech has also been widely used for data augmentation in ASR, speech translation, and related low-resource speech tasks. Prior studies show that TTS-generated audio can improve ASR by pairing available text with synthesized speech, and that TTS or voice conversion can help construct multi-speaker datasets when recording real speakers is impractical~\cite{zevallos2022data,casanova2024tts,mi2022improving}. In SQA, augmentation of spoken questions can improve robustness to noise and accent variation~\cite{sidiropoulos2022impact}. However, most augmentation work is designed for ASR, retrieval, or translation, rather than for constructing spoken question-answer pairs to train projector-based SQA models. TTS-based augmentation also supports reproducible resource creation: even when raw speech cannot be released, the pipeline, quality checks, and design choices can be shared to guide similar low-resource SQA datasets.

Important empirical findings from previous work can be clustered as follows.

\textbf{Synthetic speech quality and diversity.} Synthetic speech can produce useful SQA training data, but model gains depend strongly on the realism and diversity of the TTS voices and on whether the subsequent evaluation uses real human speech~\cite{wu2023heysquad,menevse2022framework}.

\textbf{ASR and pipeline robustness.} ASR/transcription errors substantially degrade retrieval and QA performance in cascade pipelines; end-to-end or ASR-free approaches (speech retrieval, direct speech→answer models, or projector-based speech-LLMs) are more robust when ASR is poor~\cite{lin2024speechdpr,sidiropoulos2022impact}.

\textbf{Projector-based speech→LLM systems.} Small linear projectors and LoRA-style adapters achieve strong transfer when trained on paired speech→text instruction data, but robustness to domain shift and noisy speech still requires either (a) larger/more diverse synthetic corpora, (b) pretraining the projector on high-resource languages, or (c) adapter mixtures / Mixture of Experts (MoE) techniques for multilinguality~\cite{ma2024slam}.

\textbf{Supervised projector training.} In our setup, the speech$\rightarrow$LLM connector (linear projector) is trained with supervision on paired spoken-question/text-answer data; we do not use data distillation or self-supervised objectives~\cite{ma2024slam,wang2023slm}.


In this paper, we explore this setup for Luxembourgish SQA. We start from existing text-based QA resources and an English SQA corpus, and construct a set of Luxembourgish spoken questions by combining machine translation with multiple TTS systems. We train a parameter-efficient SLAM-style model that connects a frozen Whisper large v3 turbo encoder to multilingual LLM backends via a linear projector and LoRA adapters. The model is trained to map spoken questions directly to textual answers, so that it operates in a speech instruction following regime rather than pure transcription.

\subsection{Positioning of our work}

Compared to end-to-end SQA models such as GSQA or Spectron, our approach keeps both the speech encoder and a multilingual LLM frozen and trains only a compact projector together with LoRA adapters, providing a parameter-efficient pathway from speech to textual answers~\cite{shih2024gsqa}. Unlike LLM-based ASR frameworks that optimize the projector for transcription, we directly supervise it with QA pairs so that the model learns to map spoken questions in a low-resource language to answers in text form. At the data level, we follow the spirit of TTS-based augmentation for low-resource speech, but we apply TTS to generate spoken questions from large text corpora and pair them with textual answers, rather than synthesizing generic read speech or ASR-style utterances~\cite{zevallos2022data}. 

The main contribution of this paper is therefore methodological. We present a complete workflow for constructing and exploiting TTS-augmented SQA data in a low-resource language, including translation choices, TTS selection, corpus filtering, and evaluation. Luxembourgish serves as a concrete case study, but the design and error analysis are meant to travel to other settings. We emphasize what works, what does not work, and where synthetic data falls short of human recordings, so that future data collection efforts can better balance manual and synthetic speech.


\begin{figure}[hpt!]
\centering
\includegraphics[width=0.494\textwidth]{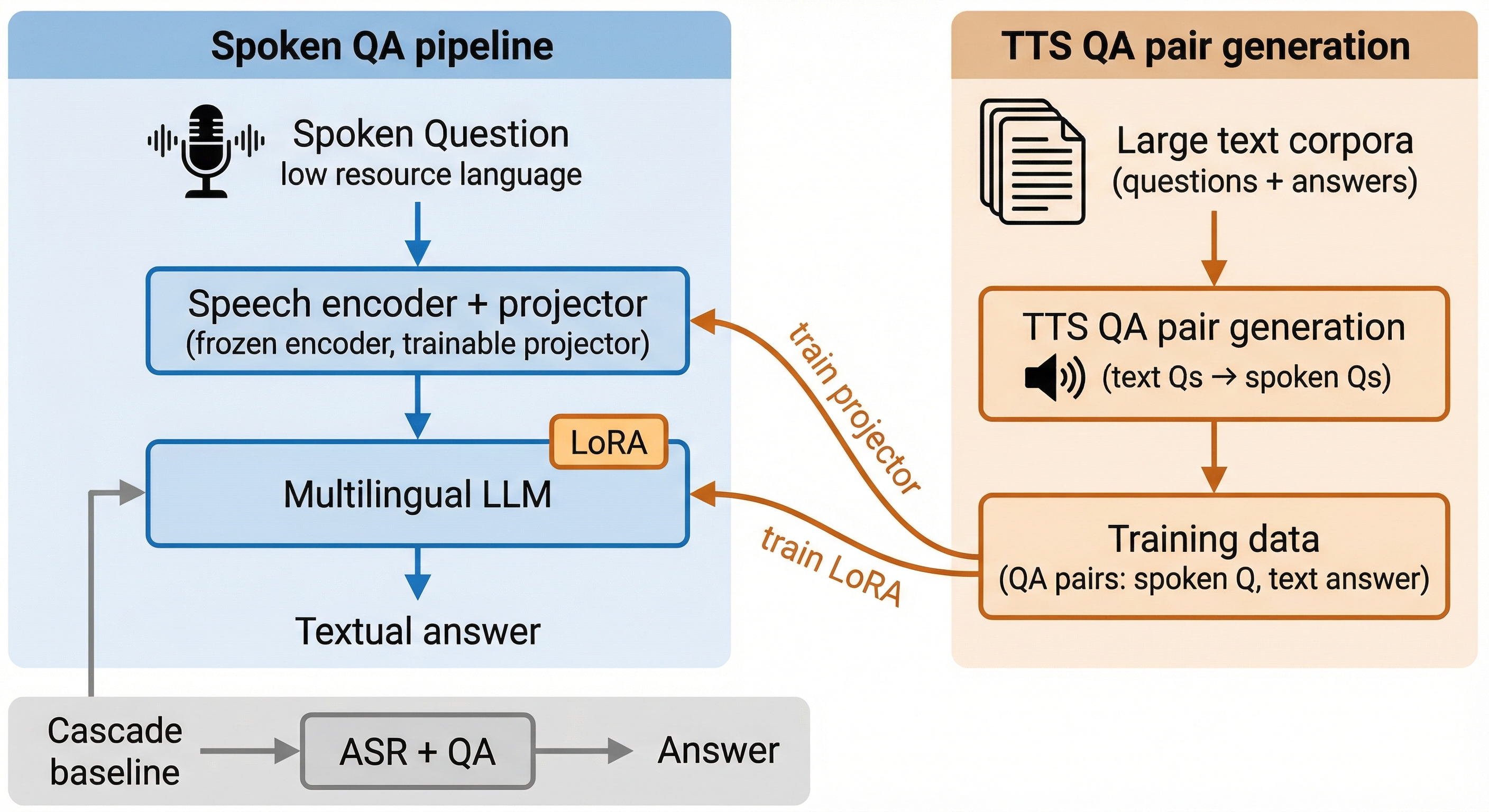}
\caption{LuxSQA training pipeline with a frozen speech encoder, a trainable projector, and a LoRA-adapted multilingual LLM backend.} 
\label{fig:training_pipeline}
\end{figure}

\section{Methodology}

\subsection{Design considerations}

We distill empirical findings from previous SQA and speech LLM work to guide the Luxembourgish pipeline and model setup (see Figure~\ref{fig:training_pipeline}). The main considerations are:
\begin{enumerate}
\item \textbf{TTS quality and naturalness.} Previous work shows that synthetic speech can provide useful training data for SQA and related tasks, but gains depend strongly on the realism of generated audio and the presence of artefacts~\cite{menevse2022framework,wu2023heysquad}. When TTS voices are more natural and closer to human prosody, models transfer better to real spoken queries. This motivates the use of high-quality TTS systems and quality control of the generated audio.
\item \textbf{Voice diversity.} Analysis of synthetic corpora such as HeySQuAD and LibriSQA indicates that using multiple voices (different speakers, genders, or timbres) reduces overfitting to a single synthetic voice and improves robustness to unseen speakers~\cite{wu2023heysquad,zhao2023librisqa}. We therefore employ several TTS engines and, where possible, multiple voices per engine, so that the synthetic Luxembourgish questions cover a range of speaking styles.
\item \textbf{Corpus size and transfer.} Projector-based systems benefit from large paired speech-text corpora, and previous work on SLAM-style architectures shows that reusing strong multilingual encoders and LLMs can improve sample efficiency for low-resource targets~\cite{fong2025speech,ma2024slam}. Rather than pretraining a new encoder, we adopt a frozen Whisper encoder and frozen EuroLLM/Apertus backends and vary the amount of synthetic and human-spoken data used to train the projector and LoRA adapters. 
\item \textbf{Pipeline robustness.} Studies on SQA and speech retrieval confirm that cascade pipelines suffer when ASR quality is low, as transcription errors propagate into the QA model~\cite{sidiropoulos2022impact,lin2024speechdpr}. Direct speech-to-response architectures, where a projector feeds a frozen LLM, are more robust under such conditions. This evidence motivated our choice of a SLAM-style speech instruction setup instead of a cascade ASR plus text QA pipeline.
\item \textbf{Generalizability of tasks.} SQA is representative of a broader class of speech-centric applications such as spoken dialogue systems, voice assistants, and audio-based reasoning; consequently, the insights gained from analyzing SQA can generalize to a wide range of other multimodal speech understanding tasks~\cite{you-etal-2022-end} with LMMs.
\end{enumerate}

These considerations shape our TTS choices, the use of multiple voices and engines, the corpus scaling experiments on synthetic versus human speech, and the decision to adopt a projector-based speech-to-answer model.

\subsection{TTS models}
\label{sec:tts}

We generate synthetic data using these models:
\begin{enumerate*}[label=(\roman*)]
\item MMS-TTS~\cite{meta-mms}. Meta's Massively Multilingual Speech TTS model is a multilingual speech synthesis system capable of generating natural speech in over a thousand languages using a unified architecture, providing consistent and scalable speech generation for multilingual evaluation. 
\item Qwen3-TTS~\cite{Qwen3-TTS}. It is a neural text-to-speech system developed within the Qwen ecosystem, designed to generate high-quality, natural-sounding speech from text with strong controllability and robustness across diverse languages; it supports wo complementary synthesis modes: CustomVoice (CV) clones a target speaker from a short reference utterance, whereas VoiceDesign (VD) generates a new voice from textual descriptions of its desired characteristics, providing controllable speech synthesis without reference audio.
\item OmniVoice~\cite{zhu2026omnivoice}, a recent state-of-the-art massively multilingual zero-shot TTS model supporting more than 600 languages. Built on a novel diffusion-based architecture inspired by language modeling, it enables high-quality speech synthesis with fast inference. Similarly to Qwen3-TTS, the model supports voice cloning (C) and controllable voice design (VD).
\end{enumerate*}

Overall, we evaluate five TTS-based synthetic data generation strategies: MMS-TTS, two Qwen3-TTS variants (Qwen-CV and Qwen-VD), and two OmniVoice variants (Omni-C and Omni-VD) as listed in Table~\ref{tab:tts_quality}.

\subsection{Data pipeline and corpora}
\label{sec:datasets}

We derive all SQA data used in this work from existing text-based QA resources. For English, we rely on SQuAD v1.1 and its spoken extension HeySQuAD~\cite{wu2023heysquad}. SQuAD consists of questions created from Wikipedia articles, where each answer is a text span extracted directly from the corresponding passage; it is considered a standardized benchmark for extractive question answering. 
HeySQuAD extends the SQuAD benchmark to SQA by introducing speech as the input modality; as a result, it is useful for developing and benchmarking voice-based question-answering systems operating under realistic acoustic conditions.
For Luxembourgish, we create synthetic speech by translating English questions into Luxembourgish using TranslateGemma 12B Instruct
model, then generating spoken questions with several neural TTS engines. Table~\ref{tab:data_overview} summarizes the resulting corpora and their roles.

\begin{table}[!t]
\centering
\scriptsize
\setlength{\tabcolsep}{2.75pt}
\renewcommand{\arraystretch}{1.1}
\caption{Speech datasets used for training and evaluation.}
\label{tab:data_overview}
\begin{tabularx}{\columnwidth}{@{}Xllll@{}}
\toprule
\rowcolor{HeaderGray}
\textbf{Name} & \textbf{Base} & \textbf{Q/S/A} & \textbf{Size} & \textbf{Use} \\
\midrule

SQuAD-LB-TTS  & SQuAD v1.1  & LB / synth LB / LB   & 48k & SQA train \\
LuxASR-50K    & LuxASR~\cite{gilles2023lux} & LB / LB / LB & 50k  & ASR train  \\ 
LuxASR-TTS-50K    & LuxASR & LB / synth LB / LB & 50k & ASR train  \\ \hline
HeySQuAD~\cite{wu2024heysquad}  & SQuAD v1.1  & EN / EN / EN    & 1002 & SQA test \\
LLAMA-EN-Test  & LLAMA test  & EN / EN / EN   & 300 & SQA test  \\
LLAMA-EN-TTS  & LLAMA test  & EN / synth EN / EN & 300  & SQA test  \\
LLAMA-LB-Test & LLAMA test  & LB / spk1--2 LB / LB & 300 & SQA test \\
\bottomrule
\end{tabularx}
\begin{tablenotes}[flushleft]
\footnotesize
\item For each resource, we indicate the source corpus, the language and modality of the question (Q), speech (S), and answer (A), the corpus size, and its role in the experimental pipeline. 
\end{tablenotes}
\end{table}

SQuAD-LB-TTS is created by translating English questions into Luxembourgish and synthesizing spoken Luxembourgish questions with the TTS engines listed above, while keeping the textual answers unchanged. 
For evaluation, we additionally use LLAMA-LB-Test, a 300-question set~\footnote{\url{https://github.com/google-research-datasets/LLAMA1-Test-Set}} whose Luxembourgish questions were human-translated and recorded by two native speakers, yielding two real-speech evaluation conditions, denoted Speaker 1 and Speaker 2 in the results.
The proprietary Luxembourgish QA set contains internal LB and EN question-answer pairs and is used as an in-domain development resource~\footnote{
\url{https://anonymous.4open.science/r/LuxSQA-samples-18A4/}}.

\subsection{TTS system quality assessment}
\label{sec:tts_quality}

To select and validate the TTS systems used for Luxembourgish question synthesis, we conducted an automatic audio quality assessment using two no-reference toolkits: 
NISQA-TTS~\cite{mittag2020deep}, which estimates naturalness MOS for synthesized 
speech, and DNSMOS~\cite{reddy2021dnsmos}, which reports overall quality 
(OVRL), speech signal quality (SIG) and background noise (BAK) on a 1--5 scale. 
We evaluated five TTS systems on 48,848 synthesized Luxembourgish utterances from the SQuAD Gemma training set. Results are reported in Table~\ref{tab:tts_quality}. Omni-C achieves the highest overall DNSMOS score (3.38), while 
Omni-VD obtains the best P.808 MOS (3.94) and is the only system 
for which NISQA-TTS naturalness was computed (3.94). Omni-VD obtains the highest P.808 MOS (3.94), followed closely by MMS-TTS (3.93), while Qwen-VD reaches 3.79.

\begin{table}[!t]
\centering
\caption{Automatic TTS quality scores.}
\label{tab:tts_quality}
\scriptsize
\setlength{\tabcolsep}{13pt}
\renewcommand{\arraystretch}{0.6}
\begin{threeparttable}
\begin{tabular}{@{}lccccc@{}}
\toprule
\rowcolor{HeaderGray}
\textbf{TTS system} & \textbf{NISQA} & \textbf{OVRL} & \textbf{SIG} & \textbf{BAK} & \textbf{P.808} \\
\midrule
MMS-TTS               & 4.15          & 3.31 & 3.56 & \textbf{4.11} & 3.93 \\
Qwen-CV      & 4.04          & 3.28 & 3.58 & 4.05 & 3.70 \\
Qwen-VD      & \textbf{4.21} & 3.33 & 3.59 & 4.09 & 3.79 \\
Omni-C       & 2.94          & \textbf{3.38} & \textbf{3.68} & 4.07 & 3.49 \\
Omni-VD          & 3.94          & 3.19 & 3.43 & 4.04 & \textbf{3.94} \\
\bottomrule
\end{tabular}
\begin{tablenotes}[flushleft]
\footnotesize
\item All scores are on a 1--5 MOS scale, where higher is better. NISQA estimates synthetic-speech naturalness. OVRL, SIG, and BAK are DNSMOS overall, speech-signal, and background-quality scores. P.808 denotes predicted listener MOS. Bold marks the best system per metric.
\end{tablenotes}
\end{threeparttable}
\end{table}

All systems achieve strong background noise scores (BAK $\geq$ 4.04), 
indicating clean synthesis without significant background artefacts. Qwen-CustomVoice (CV) scores lowest overall (DNSMOS-OVRL: 3.28), which is consistent with its lower 
downstream QA performance observed in Section~\ref{sec:results}.

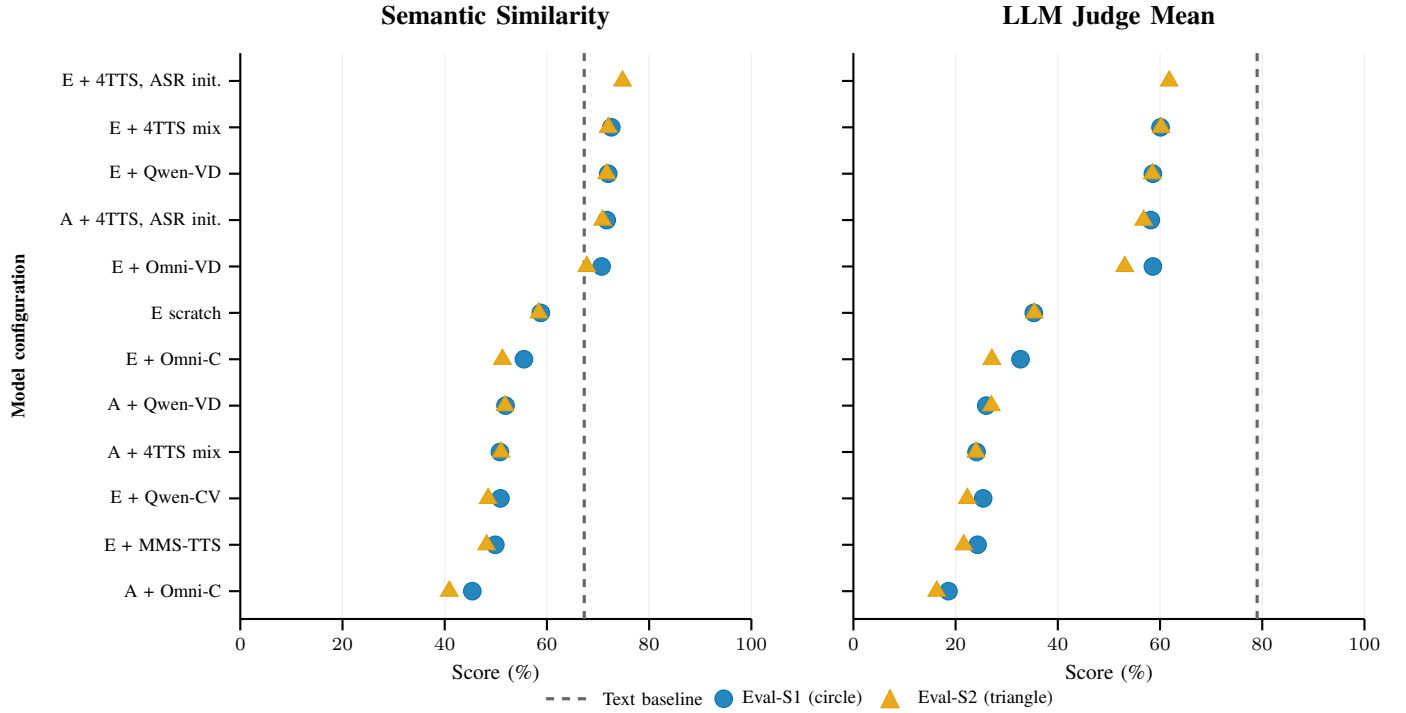
\begin{figure*}[!t]
\centering
\footnotesize
\begin{tikzpicture}

\begin{groupplot}[
    group style={
        group size=2 by 1,
        horizontal sep=1.35cm
    },
    width=0.46\textwidth,
    height=0.50\textwidth,
    xmin=0, xmax=100,
    xtick={0,20,40,60,80,100},
    xlabel={Score (\%)},
    xmajorgrids,
    grid style={gray!15},
    axis x line*=bottom,
    axis y line*=left,
    axis line style={black, thick},
    tick style={black, thick},
    tick align=outside,
    tick label style={font=\scriptsize},
    title style={font=\bfseries\normalsize},
    ylabel style={font=\bfseries\scriptsize},
    ytick={1,...,12},
    ymin=0.4, ymax=12.6,
    yticklabel style={font=\scriptsize, align=right, text width=2.36cm}
]

\nextgroupplot[
    title={Semantic Similarity},
    ylabel={Model configuration},
    yticklabels={
        {A + Omni-C},
        {E + MMS-TTS},
        {E + Qwen-CV},
        {A + 4TTS mix},
        {A + Qwen-VD},
        {E + Omni-C},
        {E scratch},
        {E + Omni-VD},
        {A + 4TTS, ASR init.},
        {E + Qwen-VD},
        {E + 4TTS mix},
        {E + 4TTS, ASR init.}
    }
]

\addplot[
    BaseGray,
    very thick,
    dashed,
    mark=none
] coordinates {
    (67.3,0.4)
    (67.3,12.6)
};

\addplot[
    only marks,
    mark=*,
    mark size=3.3pt,
    mark options={
        draw=OIblue,
        fill=OIblue!85,
        line width=0.4pt
    }
] coordinates {
    (45.4,1)
    (49.9,2)
    (50.9,3)
    (50.8,4)
    (51.9,5)
    (55.5,6)
    (58.8,7)
    (70.7,8)
    (71.7,9)
    (72.0,10)
    (72.6,11)
};

\addplot[
    only marks,
    mark=triangle*,
    mark size=3.8pt,
    mark options={
        draw=OIorange,
        fill=OIorange!90,
        line width=0.4pt
    }
] coordinates {
    (40.9,1)
    (48.2,2)
    (48.5,3)
    (51.0,4)
    (51.8,5)
    (51.3,6)
    (58.4,7)
    (67.8,8)
    (70.9,9)
    (71.7,10)
    (72.0,11)
    (74.8,12)
};

\nextgroupplot[
    title={LLM Judge Mean},
    yticklabels={,,,,,,,,,,,,}
]

\addplot[
    BaseGray,
    very thick,
    dashed,
    mark=none
] coordinates {
    (79.0,0.4)
    (79.0,12.6)
};

\addplot[
    only marks,
    mark=*,
    mark size=3.3pt,
    mark options={
        draw=OIblue,
        fill=OIblue!85,
        line width=0.4pt
    }
] coordinates {
    (18.6,1)
    (24.3,2)
    (25.4,3)
    (24.1,4)
    (26.0,5)
    (32.7,6)
    (35.3,7)
    (58.6,8)
    (58.2,9)
    (58.6,10)
    (60.1,11)
};

\addplot[
    only marks,
    mark=triangle*,
    mark size=3.8pt,
    mark options={
        draw=OIorange,
        fill=OIorange!90,
        line width=0.4pt
    }
] coordinates {
    (16.3,1)
    (21.6,2)
    (22.3,3)
    (24.0,4)
    (27.0,5)
    (27.1,6)
    (35.4,7)
    (53.1,8)
    (56.8,9)
    (58.5,10)
    (60.2,11)
    (61.8,12)
};

\end{groupplot}

\path (group c1r1.south west) -- (group c2r1.south east)
coordinate[midway] (legendmid);

\node[anchor=north, yshift=-0.75cm] at (legendmid) {
\begin{tikzpicture}
    \draw[BaseGray, very thick, dashed] (0,0) -- (0.55,0);
    \node[anchor=west, font=\scriptsize] at (0.65,0) {Text baseline};

    \filldraw[
        draw=OIblue,
        fill=OIblue!85,
        line width=0.4pt
    ] (2.35,0) circle (3.3pt);
    \node[anchor=west, font=\scriptsize] at (2.50,0) {Eval-S1 (circle)};

    \filldraw[
        draw=OIorange,
        fill=OIorange!90,
        line width=0.4pt
    ] (4.55,0.12) -- (4.43,-0.12) -- (4.67,-0.12) -- cycle;
    \node[anchor=west, font=\scriptsize] at (4.82,0) {Eval-S2 (triangle)};
\end{tikzpicture}
};

\end{tikzpicture}

\vspace{0.01cm}

\caption{Speaker-conditioned SQA performance on LLAMA-LB-Test. Blue circles denote Eval-S1, orange triangles denote Eval-S2, and dashed vertical lines mark the EuroLLM-9B text-only baseline. Points show Semantic Similarity and LLM Judge Mean for each model under two evaluation speakers. E = EuroLLM, A = Apertus, C = OmniVoice-Clone, VD = VoiceDesign, CV = CustomVoice, 4TTS mix = combined OmniVoice and Qwen-TTS data, ASR init. = initialization from an ASR-adapted checkpoint.}
\label{fig:qa_speaker_conditioned}
\end{figure*}

\subsection{Model architecture and training}

The model architecture follows the SLAM-ASR recipe~\cite{ma2024slam} and, instead of speech recognition, it is trained to answer the spoken question. We adopt \emph{Whisper-large-v3-turbo} as the audio encoder and two multilingual backend LLMs, \emph{EuroLLM 9B Instruct}~\cite{EUROLLM} and \emph{Apertus 8B Instruct}~\cite{apertus2025apertusdemocratizingopencompliant}. The speech encoder is kept frozen. Its output is a sequence of frame-level embeddings, which we downsample by a factor of $k = 5$ to reduce the length mismatch between speech and text representations. A linear projector maps the downsampled speech embeddings to the token embedding space of the chosen EuroLLM backend. The LLM is equipped with a LoRA adapter so that only the projector and LoRA parameters are updated during training. 

To study the effect of synthetic training data, we trained a family of SQA models that differ in the speech data and initialization used for projector training. We use E to denote the EuroLLM-9B-Instruct backbone and A to denote the Apertus-8B-Instruct backbone. Single-source configurations are named by the synthetic speech corpus used for projector training: MMS-TTS, Omni-C, Omni-VD, Qwen-CV, or Qwen-VD. The 4TTS mix combines the OmniVoice and Qwen3-TTS variants into a larger multi-speaker corpus of approximately 230k synthesized Luxembourgish questions, compared with about 48k questions per single-source corpus. E scratch denotes EuroLLM training without synthetic-speech initialization and ASR init. denotes initialization from an ASR-adapted checkpoint.

For each training example, the projected speech embeddings are concatenated with a short text prompt (for example, ``Answer the spoken question'') and fed to the LLM, which generates the answer tokens autoregressively. The model is optimized using cross-entropy loss over the answer tokens. 
The models were trained using the Adam optimizer with an initial learning rate of $10^{-4}$, a linear warmup of 1,000 steps, and a subsequent scheduled learning rate decay. Training was carried out for 5 epochs with a batch size of 16. LoRA is applied with rank $r = 16$ and scaling factor $\alpha = 16$.
At inference time, only the spoken question and prompt are provided, and the model returns a textual answer.

\subsection{Experimental conditions and metrics}

We structure our experiments around three research questions: \begin{enumerate*}[label=(\roman*)]
\item[(a)] \textbf{Synthetic data and initialization.} How do TTS-generated Luxembourgish QA data and ASR-adapted initialization affect downstream SQA performance? We compare single-source TTS corpora, a multi-source 4TTS mix, training from scratch, and initialization from an ASR-adapted checkpoint.
\item[(b)] \textbf{TTS source and voice diversity.} How does the choice of TTS source and the use of multi-source synthetic speech affect robustness across evaluation speakers? We compare models trained on MMS-TTS, Qwen3-TTS, OmniVoice, and the combined 4TTS mix.
\item[(c)] \textbf{Evaluation and transfer.} How well do semantics-aware metrics capture Luxembourgish SQA performance on LLAMA-LB-Test, and what do the results indicate about the transferability of TTS-augmented training to real spoken questions?
\end{enumerate*}


Because the model generates answers autoregressively in a morphologically rich,
low-resource language where predictions paraphrase the gold answer or
code-switch between Luxembourgish, German, French, and English, exact-match and
Token-overlap metrics are too brittle to be informative. We evaluate with two semantics-aware metrics, reported as percentages. \textbf{Semantic Similarity (SemSim)} embeds the gold answer and prediction
with a multilingual sentence encoder (\texttt{paraphrase-multilingual-MiniLM-L12-v2}) and computes the maximum cosine
similarity between the gold answer and any prediction span; this rewards predictions containing the correct answer even within longer or repetitive
output. 
\textbf{LLM Judge Mean} uses two instruction-tuned judges, GPT-4o and
Gemini-2.5-Flash, run at temperature~0. Each judge assigns a score from
$\{0, 0.5, 1\}$ to every (question, gold answer, prediction) triple under a
fixed partial-credit rubric that accepts translations, aliases, and spelling or
ASR variants of the gold answer. We average the two judge scores per item and
report the mean of these item-level averages over the test set. The two metrics
are independent by construction, one embedding-based and one LLM-based, so
their agreement on model ranking serves as a robustness check.

\section{Experimental results}
\label{sec:results}

We report results on the LLAMA1-Test-Set using both text-only baselines and spoken Luxembourgish input. Figure~\ref{fig:qa_speaker_conditioned} summarizes the main comparison in terms of Semantic Similarity and LLM Judge Mean. These results address three questions~\ref{sec:datasets}: the usefulness of synthetic data and ASR-adapted initialization for low-resource SQA, the effect of TTS source and voice diversity across evaluation speakers, and the extent to which semantics-aware metrics and automatic TTS quality estimates explain downstream SQA performance.


\subsection{Overall QA performance}



The text-only baselines provide an upper bound once the acoustic channel is removed. EuroLLM-9B (base) reaches 67.3 SemSim and 79.0 LLM Judge Mean on English text input, while the non-LB-adapted text baselines score lower in SemSim: 51.6 for EuroLLM-9B and 51.3 for Apertus-8B. This shows that strong text-side generation alone is not sufficient for SQA.

Luxembourgish audio evaluation is substantially harder without speaker-specific synthetic training. Generic lB-adapted models reach 45.7 SemSim and 62.5 LLM Judge Mean for Apertus-8B, and 38.8 SemSim and 32.8 LLM Judge Mean for EuroLLM-9B, showing that generic Luxembourgish adaptation does not close the speech-text gap.

\subsection{Effect of TTS choice and speaker-specific training}

The strongest results are obtained by models trained on speaker-specific synthetic Luxembourgish speech. For Speaker~1, the best system is EuroLLM-9B (4tts), with 72.6 SemSim and 60.1 LLM Judge Mean. For Speaker~2, the best system is EuroLLM-9B (4tts-from-asr), with 74.8 SemSim and 61.8 LLM Judge Mean. These scores are not only far above the generic \texttt{\_lb} models, but also close to or above the semantic similarity of the text baseline. This indicates that carefully constructed synthetic training data can compensate for the lack of real Luxembourgish QA recordings. The comparison between A + 4TTS mix and A + 4TTS, ASR init. Further shows that ASR initialization helps: using an ASR-adapted checkpoint trained with both original LuxASR and synthesized LuxASR-TTS data raises the result on both speakers.

Across both speakers, EuroLLM is stronger than Apertus in higher-performing configurations. Qwen-VD, Omni-VD, and 4tts yield the most competitive systems, while Qwen-CV, MMS-TTS, and Omni-C tend to lag. The gap is especially clear in the LLM Judge Mean, where weaker TTS configurations often fall to the low 20s or 30s despite moderate semantic similarity. This suggests that some systems preserve enough lexical content for partial semantic matching but still produce responses judged less appropriate or less well-formed.

The Speaker~1 and Speaker~2 rankings are also broadly stable. High-performing configurations remain near the top for both speakers, while weaker configurations remain weak across both test conditions. This stability strengthens the conclusion that the main driver is the training data configuration rather than speaker-specific randomness in the test set.

\subsection{Relation between TTS quality and downstream QA}
Table~\ref{tab:tts_quality} shows that the synthesized speech is clean, with uniformly strong DNSMOS background scores and relatively narrow ranges in no-reference quality estimates. However, downstream QA performance does not follow these perceptual metrics monotonically. For example, Omni-C achieves the highest DNSMOS overall score, yet its downstream QA results are below those of the strongest 4tts and voice-design configurations. Similarly, MMS-TTS receives strong perceptual scores but remains among the weaker setups in QA.

This mismatch indicates that good TTS quality is necessary but not sufficient for effective SQA training. What matters for downstream performance is not only naturalness but also whether the synthetic speech preserves the lexical, phonetic, and prosodic cues that help the speech encoder and projector recover the question content reliably. In our setting, speaker-aware and multi-source synthetic datasets appear more useful than choosing one TTS system based only on MOS-like scores.

\section{Discussion}
Our results support three conclusions. First, TTS-based augmentation is a viable strategy for low-resource SQA. This is consistent with previous SQA work showing that SQA data can be constructed from text-based QA resources using TTS and ASR components, and with HeySQuAD, which extends SQuAD with human-spoken and machine-generated questions~\cite{menevse2022framework,wu2023heysquad}. It also aligns with the work on data scarcity in SQA, where automatically generated question-answer pairs are used to improve QA performance under limited annotated data conditions~\cite{menevcse2024dealing}. In our Luxembourgish setting, generic audio models remain weak, but speaker-specific synthetic training substantially improves performance on LLAMA-LB-Test. This is an important practical finding because it shows that a usable SQA pipeline can be built without collecting a large human-recorded Luxembourgish QA corpus.

Second, the choice of synthetic training data matters more than the perceptual quality of TTS. Previous work on TTS model selection for synthetic data generation has shown that computable TTS quality metrics such as NISQA MOS and intelligibility do not necessarily predict downstream ASR performance~\cite{rossenbach2024problem}. Our results show a similar pattern for SQA: the best downstream models are obtained from 4tts and voice-design-based setups, rather than from systems with the strongest DNSMOS or MOS estimates. This suggests that downstream SQA should evaluate TTS not only as speech synthesis but also as a task-specific data generation mechanism.

Third, the results highlight a persistent gap between text and speech. Previous work has shown that SQA and spoken retrieval are affected by ASR errors, acoustic variability, and speech-text mismatch~\cite{sidiropoulos2022impact,lin2024speechdpr}. In our experiments, even when semantic similarity is high for the best speaker-specific systems, LLM Judge Mean remains below the text-only baseline. This implies that the spoken pipeline still loses information relative to text input, likely through a combination of translation noise, pronunciation mismatches, synthetic-speaker bias, and imperfect speech-to-LLM alignment. The current approach is therefore effective, but it does not remove the need for better Luxembourgish speech coverage, more diverse speakers, and stronger evaluation on naturally recorded audio.

Taken together, these findings position LuxSQA as a data-centric recipe rather than a model comparison alone. The main contribution is that it identifies which types of synthetic speech are most useful for low-resource SQA and shows that carefully designed TTS augmentation can produce meaningful gains when real SQA data are scarce.

\subsection{Limitations and Future Work}
Although experiments show that synthetic data can support SQA in a low-resource language, several aspects require controlled evaluation. First, we compare different training configurations, but do not isolate the effect of data scale. Future work should measure how SQA performance changes with synthetic dataset size, TTS engine, and language coverage.

Second, the current datasets rely on a limited number of synthetic speakers. Prior SQA datasets such as HeySQuAD show that human-spoken and machine-generated questions differ in noise profile and evaluation behavior~\cite{wu2023heysquad}. A systematic speaker-diversity study should therefore vary the number of voices, speaker characteristics, and speaking styles to test robustness to unseen speakers.

Third, our models map spoken questions directly to textual answers without explicit access to supporting context. This design isolates speech understanding, but many real-world QA systems require retrieval or reasoning over external evidence. Future work should examine context-aware SQA, where the model jointly processes the spoken question and associated documents or facts, following the broader motivation of open-domain SQA and spoken passage retrieval~\cite{sidiropoulos2022impact,lin2024speechdpr}.

In addition, LLM Judge Mean inherits the known self-preference, verbosity, and reproducibility limitations of LLM-as-judge evaluation~\cite{zheng2023judging,liu2023g}; we reduce these with two independent judges at temperature~0 and a fixed rubric, and corroborate results with SemSim, but human evaluation remains future work.

Finally, Luxembourgish is one low-resource case study. Additional experiments on languages with different phonological and morphological properties are needed to test whether the proposed pipeline generalizes. Evaluation of naturally recorded speech from multiple speakers and acoustic conditions would further clarify robustness and transferability.

\section{Conclusions}

This paper presented LuxSQA, a workflow for constructing Luxembourgish SQA data from text-based QA resources using machine translation and multilingual TTS. Using a SLAM-style architecture with a frozen Whisper encoder, a learned projector, and LoRA-adapted multilingual LLMs, we showed that synthetic Luxembourgish speech can improve spoken QA performance when real QA recordings are scarce. The strongest results came from multi-source and speaker-aware TTS training configurations, while no-reference TTS quality scores alone did not reliably predict downstream QA performance. These findings suggest that TTS augmentation is useful for low-resource SQA, but that synthetic data should be evaluated as task-specific training data, not only as natural-sounding speech.

\section*{Generative AI Disclosure}
Generative AI (Grammarly) tools were used to support English language editing and stylistic rephrasing. The scientific content, analyses, and conclusions remain the responsibility of the authors.

\bibliographystyle{IEEEtran}
\bibliography{mybib}
\end{document}